\documentclass{article}

\usepackage{microtype}
\usepackage{graphicx}
\usepackage{subfigure}
\usepackage{booktabs} 

\usepackage{hyperref}
\usepackage{subfiles}
\usepackage{multirow}
\usepackage{amsmath}
\usepackage{subcaption}
\usepackage{mathtools}
\usepackage{booktabs}

\usepackage{amsthm}
\usepackage{nicematrix}
\usepackage{amsfonts}
\usepackage{gensymb}
\usepackage{tikz}

\usepackage[accepted]{icml2025}

\usepackage{amsmath}
\usepackage{amssymb}
\usepackage{mathtools}
\usepackage{amsthm}

\usepackage{algorithm}
\usepackage{algorithmic}
\usepackage[capitalize,noabbrev]{cleveref}

\usepackage[textsize=tiny]{todonotes}

\icmltitlerunning{Hierarchical Motion Captioning Utilizing External Text Data Source}

\begin{document}

\twocolumn[
\icmltitle{Hierarchical Motion Captioning Utilizing External Text Data Source}



\icmlsetsymbol{equal}{*}

\begin{icmlauthorlist}
\icmlauthor{Clayton Leite}{yyyd}
\icmlauthor{Yu Xiao}{yyy}
\end{icmlauthorlist}
\icmlaffiliation{yyy}{Aalto University, Espoo, Finland}

\icmlcorrespondingauthor{Yu Xiao}{yu.xiao@aalto.fi}


\vskip 0.3in
]






\begin{abstract}

This paper introduces a novel approach to enhance existing motion captioning methods, which directly map representations of movement to high-level descriptive captions (e.g., ``a person doing jumping jacks").  The existing methods require motion data annotated with high-level descriptions (e.g., ``jumping jacks"). However, such data is rarely available in existing motion-text datasets, which additionally do not include low-level motion descriptions.
To address this, we propose a two-step hierarchical approach. First, we employ large language models to create detailed descriptions corresponding to each high-level caption that appears in the motion-text datasets (e.g., ``jumping while synchronizing arm extensions with the opening and closing of legs" for ``jumping jacks"). These refined annotations are used to retrain motion-to-text models to produce captions with low-level details. Second, we introduce a pioneering retrieval-based mechanism. It aligns the detailed low-level captions with candidate high-level captions from additional text data sources, and combine them with motion features to fabricate precise high-level captions. Our methodology is distinctive in its ability to harness knowledge from external text sources to greatly increase motion captioning accuracy, especially for movements not covered in existing motion-text datasets. Experiments on three distinct motion-text datasets (HumanML3D, KIT, and BOTH57M) demonstrate that our method achieves an improvement in average performance (across BLEU-1, BLEU-4, CIDEr, and ROUGE-L) ranging from 6\% to 50\% compared to the state-of-the-art M2T-Interpretable.
\end{abstract}


\section{Introduction}

Motion captioning, also known as motion-to-text, involves generating textual descriptions of human motions based on time-series data, such as video and skeletal pose trajectories. It has a wide range of applications across various domains. In assistive technologies, it can provide people with disabilities -- e.g. visual impairments -- with detailed descriptions of movements they cannot perceive visually \cite{Safiya2024}. In smart home environments, it can help monitor physical activities of elderly people to detect potential declines in their mental or physical capabilities \cite{10.1145/3607720.3607746}. In this paper, we focus on addressing the challenge of motion captioning utilizing sequences of body joint parameters such as joint angles and angular velocities. These motion representations can be obtained from a variety of sources, such as motion caption systems and wearable inertial measurement units. 


Existing methods \cite{motiongpt, lamp, m2tinter, t2mgpt} are trained on motion-text datasets \cite{humanml3d, kitml, both57m} containing motion data annotated with high-level textual descriptions, such as ``a person doing jumping jacks.” Since these datasets, including KIT \cite{kitml}, HumanML3D \cite{humanml3d}, and BOTH57M \cite{both57m}, do not provide annotations with low-level motion descriptions, such as ``the person is jumping while synchronizing arm extensions with the opening and closing of their legs” for ``jumping jacks”, the existing methods tend to learn a direct mapping from motion data to high-level motion description. However, the performance of such methods is limited due to the scarcity of annotated examples for different motions in motion-text datasets. For example, the term ``jumping jacks” appears only once in the entire KIT dataset \cite{kitml}. Similarly, other high-level terms such as ``playing ping-pong,” ``performing gymnastics,” or ``doing judo” are equally underrepresented. In fact, half of the terms in the KIT dataset appear at most twice across the entire dataset. The straightforward solution to this challenge is to collect more data. However, recording high-quality motion data requires expensive motion capture systems. This raises a critical question: \textit{how can we improve the generation of high-level textual descriptions without relying on extensive data collection of annotated human motion data?}

Unlike previous motion captioning methods that train exclusively on paired motion and high-level caption examples, we propose harnessing additional text data sources to enhance the understanding of human motions. Just as humans can learn what it means to perform jumping jacks through a detailed explanation, such as ``jumping while synchronizing arm extensions with the opening and closing of the legs," without seeing the motion firsthand, we aim to generate more accurate high-level captions based on detailed motion descriptions, and utilize the external data sources to learn the relationship between high-level and low-level descriptions. Motivated by this analogy, we replace the one-step direct mapping with a two-step hierarchical approach named \textbf{Hi}erarchical \textbf{C}aption-\textbf{A}ugmented \textbf{M}otion-to-\textbf{T}ext (HiCAM2T).

First, we leverage large language models to transform high-level captions, such as ``performing jumping jacks,” into detailed low-level descriptions that explain the movement of body joints. This allows us to annotate all motion data in existing motion-text datasets, which previously only had high-level descriptions, with detailed low-level descriptions as well. By using a motion encoder and a text decoder, our method learns to generate low-level motion descriptions for each input motion.

Second, given these low-level descriptions, our system retrieves potential high-level descriptions that match the low-level details from a database. This database, which is part of our pipeline, is populated with pairs of high-level and low-level descriptions obtained from multiple sources, including existing motion-text datasets and external text sources. The database can be expanded with additional motion descriptions at inference time (with the computational cost increasing linearly), enabling the model to dynamically handle new high-level motion descriptions. The retrieved high-level descriptions are then combined with motion features by the text decoder to generate the final high-level motion description, ensuring coherence with the motion.

The hierarchical design of HiCAM2T eliminates the need for additional, costly motion data collection. To the best of our knowledge, this work is the first to tackle the motion captioning task by emphasizing a detailed understanding of motions and leveraging both paired motion-text data and standalone text descriptions (i.e., without corresponding motion) to enhance motion captioning performance.

We evaluate the performance of HiCAM2T using three distinct motion-text datasets: HumanML3D \cite{humanml3d}, KIT \cite{kitml}, and BOTH57M \cite{both57m}. Our results demonstrate that HiCAM2T achieves an improvement in average performance (across BLEU-1, BLEU-4, ROUGE-L, and CIDEr) ranging from 6\% to 50\% compared to state-of-the-art methods such as M2T-INT \cite{m2tinter}. Additionally, we assess the impact of enriching the database at inference time with additional motion descriptions, which were not seen during training. This enrichment results in up to a 13\% increase in average performance. Through ablation studies we also validate the effectiveness of the key elements of HiCAM2T.

The remainder of this paper is organized as follows. Section 2 reviews the related work. Section 3 provides a detailed explanation of HiCAM2T. Section 4 outlines the experimental setup, presents the results, and includes a discussion. Finally, Section 5 concludes the paper.

\section{Related Work}

Inspired by Vector Quantized Variational Autoencoders (VQ-VAE) \cite{vqvae}, \citet{chuan2022tm2t} tackled the motion-to-text task by first learning a discrete representation of motion -- this process is known as motion tokenization. This intermediate representation was then translated into words describing the motion. More specifically, the authors utilized a Transformer encoder network to map the discrete motion representations into feature vectors, which were then processed in a Transformer decoder to predict the probability distribution over discrete language tokens. 

MotionGPT \cite{motiongpt} adopted a similar approach with the distinction that the authors leveraged the pre-trained T5 language model \cite{t5} to encode discrete motion representations and decode the feature vectors into text. They fine-tuned the pre-trained T5 model on motion-text data, with both unsupervised and supervised strategies following \citet{t5}. LaMP \cite{lamp} similarly employed a discretization approach for the motion and a Transformer to encode these discrete motion representations into feature vectors. These feature vectors were then decoded into text using a pre-trained large language model (LLM), specifically OPT-2.7B \cite{opt27b}. To enhance the model's performance, the authors proposed fine-tuning the LLM using LoRA (Low-Rank Adaptation) \cite{lora}. MotionLLM \cite{motionllm} follows the same approach as LaMP.

The main challenge in motion tokenization lies in the inherent loss of information during the conversion of continuous values into discrete representations \cite{momask}. An additional challenge is codebook collapse, where only a small subset of codebook entries is utilized, thus limiting the tokenizer's capacity to capture diverse patterns. \citet{m2tinter} abstained from the motion tokenization approach. The authors introduced a novel architecture, M2T-Interpretable, specifically designed for the motion-to-text task. M2T-Interpretable directly encodes the continuous motion representation into feature vectors. This motion encoding process incorporates attention mechanisms to selectively focus on specific segments of the human body at particular moments. The motion captions are then generated by mapping these encoded features to text using an LSTM model.

\begin{figure*}[tb]
    \centering
    \includegraphics[width=1.0\linewidth]{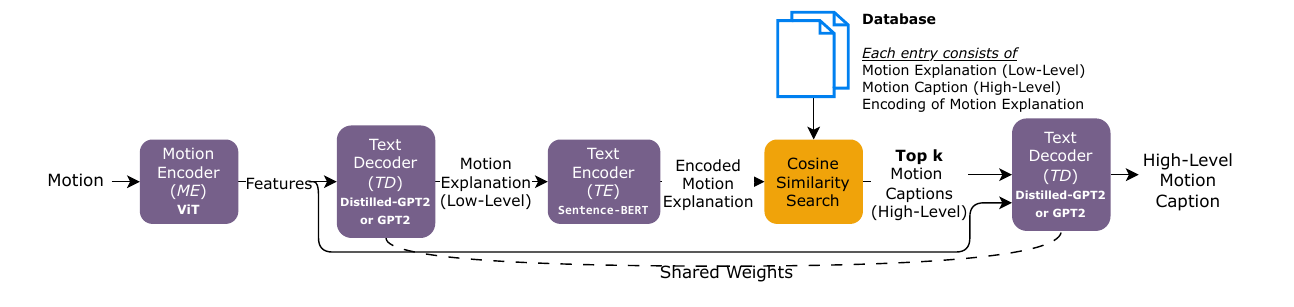}
    \vspace*{-0.5cm}
    \caption{System overview of HiCAM2T. Given a motion as input, the motion encoder (ViT) extracts relevant motion-related features, which are then passed to the text decoder to generate text encompassing a low-level description of the motion. This low-level description is then encoded and compared with existing low-level descriptions in the database using cosine similarity search. The high-level descriptions corresponding to the most similar low-level descriptions are retrieved from the database and utilized to generate a final high-level motion caption using either the Distilled-GPT2 or GPT2 model. The selection of the text decoder is treated as a hyper-parameter, chosen based on the optimization of validation performance. We choose to share the weights of the text decoders because preliminary experiments indicated no significant performance differences when using two separate text decoders.}
    \label{fig:sys_overview}
\end{figure*}

\begin{figure}[tb]
    \centering
    \includegraphics[width=1\linewidth]{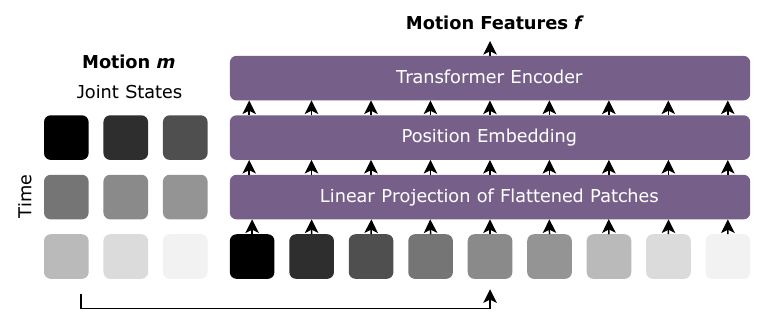}
    \caption{Vision Transformer neural network architecture utilized for the motion encoder. The motion, represented as a two-dimensional structure, is divided into equally sized rectangular patches. These patches are then linearly projected and summed with a learnable positional encoding. The resulting representation is fed through multiple Transformer layers in the encoder to generate the motion features.}
    \label{fig:vit}
    \vspace{-0.5cm}
\end{figure}

In summary, previous methods have overlooked the challenge of directly mapping motion data to high-level captions, instead focusing on designing distinct motion encoding architectures \cite{chuan2022tm2t, m2tinter} or language modeling techniques \cite{motiongpt, lamp}. In contrast, our approach HiCAM2T offers a different perspective by addressing the inherent nature and limitations of existing motion-text datasets. HiCAM2T stands out from previous methods by 1) employing a hierarchical framework that first learns low-level motion descriptions before generating high-level captions; and 2) harnessing standalone motion captions to enhance performance. Furthermore, we abstain from the use of motion tokenization to circumvent its known limitations.

\section{Methodology}

In this section, we provide an overview of HiCAM2T with a detailed explanation of each of its components.

\subsection{System Overview}

Fig. \ref{fig:sys_overview} provides an overview of our system. It includes three distinct neural networks: a motion encoder (\textit{ME}), a text decoder (\textit{TD}), and a text encoder (\textit{TE}). A motion is represented as a two-dimensional tensor $\pmb{m}$: \textit{time × joint states}, where the joint states contain body joint angles and velocities. Provided as input, the motion is processed by the motion encoder, based on a Vision Transformer (ViT) \cite{vit}, extracting feature vectors $\pmb{f}$. This is expressed in Eq. \ref{eq:features}.
\begin{equation}
\pmb f = \text{ME}_{\theta}(\pmb m) 
\label{eq:features}
\end{equation}

where $\theta$ denotes the parameters of the \textit{ME} model. These features $\pmb{f}$ are then passed to the text decoder as if they were text embeddings, which generates a low-level description of the limb and torso movements (i.e., movement explanation), expressed as $\pmb{\hat{z}}$ in Eq. \ref{eq:zhat}. Note that the choice of the text decoder, pre-trained Distilled-GPT2 \cite{sanh2019distilbert} or GPT2 \cite{radford2019language}, is a hyper-parameter to be selected to optimize the validation performance.
\begin{equation}
\pmb{\hat{z}} = \text{TD}_{\phi}(\pmb f)
\label{eq:zhat}
\end{equation}

where $\phi$ denotes the parameters of the \textit{TD} model. The parameters of the \textit{ME} and the \textit{TD} models are optimized by maximizing the log-likelihood of the probability distribution $p_\phi (z_t \mid \pmb{f})$ expressed in Eq. \ref{eq:prob_dist}. 
\begin{equation}
    p_\phi (z_t \mid \pmb{f}) = \prod_i^{T-1} p_\phi \left( z_i \mid z_{<i}, \pmb{f} \right)
    \label{eq:prob_dist}
\end{equation}

where $\pmb{z} = \{z_1, z_2, \dots, z_T\}$ is the ground-truth low-level description (represented as tokens) of the motion $m$, $T$ is the length of $\pmb{z}$, and $z_t$ is the \textit{t}-th token of $\pmb{z}$. 

In other words, the parameters $\phi$ and $\theta$ are optimized by minimizing the loss expressed in Eq. \ref{eq:l1}.
\begin{equation}
\mathcal{L}_1 = - \sum_{t=1}^T \log p_\phi(z_t \mid {z}_{<t}, \pmb f)
\label{eq:l1}
\end{equation}

The detailed explanation generated by the text encoder is processed by a text decoder, which is designed to produce embeddings that capture semantic similarity. The decoder ensures that explanations of similar motions yield embeddings with high cosine similarity. We utilize the pre-trained Sentence-BERT \cite{sentence-bert} model as the text decoder due to its alignment with our purpose. We refer to the embeddings generated by \textit{TD} as $\pmb{\hat{u}}$ (Eq. \ref{eq:embeddings_uhat}).
\begin{equation}
\pmb{\hat{u}} = \text{TE}_{\psi}(\pmb{\hat{z}})
\label{eq:embeddings_uhat}
\end{equation}

The embeddings $\pmb{\hat{u}}$ are used to retrieve from a database the top $k$ high-level motion descriptions based on their relevance to $\pmb{\hat{u}}$. Here, $k$ is a user-defined parameter that determines the number of descriptions to retrieve. This retrieval is performed by comparing $\pmb{\hat{u}}$ with the embeddings of low-level motion descriptions using the cosine similarity function. Descriptions with the highest cosine similarity scores, indicating the greatest semantic similarity to $\pmb{\hat{u}}$, are selected as the top $k$ matches.

The parameters of the text encoder, denoted as $\psi$, are fine-tuned using the contrastive loss function defined in Eq.  \ref{eq:l2}. This fine-tuning process aims to optimize the decoder for better semantic similarity matching.
\begin{equation}
    \mathcal{L}_{2} = 1 - \dfrac{\pmb{\hat{u}} \cdot \pmb{u}}{  \lvert \lvert \pmb{\hat{u}} \rvert\rvert \text{ } \lvert \lvert \pmb{u}\rvert\rvert  } + \max(0, c - \dfrac{\pmb{\hat{u}} \cdot \pmb{\bar{u}}}{  \lvert \lvert \pmb{\hat{u}} \rvert\rvert \text{ } \lvert \lvert \pmb{\bar{u}}\rvert\rvert  } )
    \label{eq:l2}
\end{equation}

where $\pmb{u}$ denotes the embeddings of the ground-truth low-level description of motion $\pmb{m}$, $\pmb{\bar{u}}$ refer to the embeddings of a low-level description of a randomly selected motion $\pmb{m'} \neq \pmb{m}$. Here, $c$ is a real-valued constant in the interval [0,1], defined by the user. This constant serves as a hyper-parameter that balances the trade-off between pulling similar embeddings closer and pushing dissimilar embeddings apart. Higher values of  $c$ (closer to 1) emphasizes pushing apart dissimilar embeddings. Note that since the retrieval process relies on cosine similarity search, its complexity scales linearly with the size of the database.

Finally, the features $\pmb f$ of the motion $\pmb m$ are concatenated with the top $k$ retrieved high-level captions. The text decoder then processes this input to generate the final high-level caption. This concatenation and subsequent processing by the text decoder is necessary for a more accurate final high-level caption, instead of simply using the top-ranked retrieved high-level caption as the final output. In the ablation studies, we analyze the impact of directly using the top-retrieved high-level caption as the final output, bypassing the processing step with the text decoder.

The loss expressed in Eq. \ref{eq:l3} is utilized to optimize the parameters of the text decoder and the motion encoder.
\begin{equation}
    \mathcal{L}_{3} =  - \sum_{t=1}^T \log p_\omega(y_t \mid {y}_{<t}, \pmb{r}^{(1)}, \pmb{r}^{(2)}, ..., \pmb{r}^{(k)},  \pmb f)
    \label{eq:l3}
\end{equation}

where $\pmb{r}^{(1)}, \pmb{r}^{(2)}, ..., \pmb{r}^{(k)}$ denotes the top $k$ retrieved high-level descriptions and $\pmb{y} = \{ y_1, y_2, \dots, y_T\}$ the ground-truth high-level description. The output of the system is denoted as $\pmb{\hat{y}}$ (Eq. \ref{eq:sys_out}).
\begin{equation}
\pmb{\hat{y}} = \text{TD}_{\phi}(\pmb{r}^{(1)}, \pmb{r}^{(2)}, ..., \pmb{r}^{(k)},  \pmb f)
\label{eq:sys_out}
\end{equation}

\begin{algorithm}[tb]
\caption{Training algorithm of our system. $N$ is the number of pairs of motion-caption in the dataset. $\mathcal{D}$ represents the database, which contains the motions, their corresponding high-level captions, low-level captions, and the encoded representations of the low-level captions.}
\begin{algorithmic}[1]
\REQUIRE $\mathcal{D} = \{(\pmb{m}_i, \pmb{y}_i, \pmb{z}_i) \text{ } | \text{ } i = 1, 2, \dots, N\}$
\REQUIRE Define $n\_epochs$, $c$, $k$, $\lambda_1$,  $\lambda_2$, and  $\lambda_3$
\STATE Initialize models ME, TD, and TE
\FOR{$\pmb{z}$ in $\mathcal{D}$}
    \STATE Encode low-level description $\pmb{u} = \text{TE}_{\psi}(\hat{z})$
    \STATE Store $\pmb{u}$ with its corresponding $\pmb{y}$ (high-level caption)
\ENDFOR
\STATE Let $\mathcal{U} = \{(\pmb{u}_i, \pmb{y_i}) \text{ } | \text{ } i = 1, 2, \dots, N\}$
\FOR{epoch in $n\_epochs$}
\FOR{each $\pmb{m}, \pmb{y}, \pmb{z}$ in $\mathcal{D}$}
    \STATE Motion encoding $\pmb f = \text{ME}_{\theta}(\pmb m)$
    \STATE Low-level description generation $\pmb{\hat{z}} = \text{TD}_{\phi}(\pmb f)$
    \STATE Compute $\mathcal{L}_1$  according to Eq. \ref{eq:l1}
    \STATE Encode low-level description $\pmb{\hat{u}} = \text{TE}_{\psi}(\pmb{\hat{z}})$
    \STATE Randomly select $\pmb{\bar{u}} \in \mathcal{U}$ from a motion $\pmb{m'} \neq \pmb{m}$
    \STATE Compute $ \mathcal{L}_{2}$  according to Eq. \ref{eq:l2}
    \STATE Let $\mathcal{V} = \{ (\pmb{v}_i = \dfrac{\pmb{\hat{u}} \cdot \pmb{u}_i}{  \lvert \lvert \pmb{\hat{u}} \rvert\rvert \text{ } \lvert \lvert \pmb{u}_i\rvert\rvert  }, \pmb{y}_i) \text{ } | \text{ } (\pmb{u}_i. \pmb{y}_i) \in \mathcal{U} \}$
    \STATE Sort $\mathcal{V} = \{ (\pmb{v}^{(i)}, \pmb{y}^{(i)}) \}$ s.t. $v^{(i)} \geq v^{(i+1)} \text{ } \forall i$
    \STATE Retrieve $\pmb{r}^{(1)}, \pmb{r}^{(2)}, ..., \pmb{r}^{(k)}$ =  $\pmb{y}^{(1)}, \pmb{y}^{(2)}, ..., \pmb{y}^{(k)}$
    \STATE Compute $\mathcal{L}_{3}$  according to Eq. \ref{eq:l3}
    \STATE Compute $\mathcal{L} = \lambda_1 \mathcal{L}_1  + \lambda_2 \mathcal{L}_2  + \lambda_3 \mathcal{L}_3 $
    \STATE Update $\theta, \phi, \psi$ according to $\nabla_{\theta, \phi, \psi} \mathcal{L}$
\ENDFOR
\FOR{$\pmb{z}$ in $\mathcal{D}$}
    \STATE Encode low-level description $\pmb{u} = \text{TE}_{\psi}(\hat{z})$
    \STATE Update $\pmb{u}$ with its corresponding $\pmb{m}, \pmb{y}, \pmb{z}$
\ENDFOR
\ENDFOR
\end{algorithmic}
\vspace{-0.5cm}
\label{alg:alg}
\end{algorithm}

The total loss function $\mathcal{L}$ is defined as the weighted sum of the three individual loss terms described earlier (Eq. \ref{eq:total_l}). Algorithm \ref{alg:alg} presents the pseudocode detailing the training process of our system.
\begin{equation}
    \mathcal{L} = \lambda_1 \mathcal{L}_1  + \lambda_2 \mathcal{L}_2  + \lambda_3 \mathcal{L}_3 
    \label{eq:total_l}
\end{equation}

\subsection{Motion Encoder}

We implemented a Vision Transformer \cite{vit} (ViT) as the neural network architecture for the motion encoder (Fig. \ref{fig:vit}). No modifications to the ViT architecture are required, as the motion can be represented as a single-channel image. Given a two-dimensional tensor representing motion, the ViT architecture extracts rectangular patches and processes each patch in parallel through a fully connected layer. The resulting patch embeddings are summed with learnable positional embeddings before being passed through the transformer layers. Within the transformer layers, the patches are projected to match the embedding dimension of the text encoder. The output features of each motion is also a two-dimensional tensor. In addition to its parallelization capabilities, the decision to use a Vision Transformer was influenced by the findings of Leite et al. \cite{leite2024transformerbasedapproachessensorbasedhuman}, which showed that this architecture excels at capturing the spatial and temporal relationships intrinsic to motion data.

\begin{table*}[tb]
\centering
\begin{tabular}{p{6.5cm}p{9.5cm}}
\hline
\textbf{Motion Caption (High-Level Description)} &
  \textbf{Motion Explanation (Low-Level Description)} \\ \hline
A human performs a boxing  punch. &
  The arms and legs move in a coordinated motion  to generate power and speed for the punch. The  arms extend forward and the legs pivot to transfer  weight to the front foot, which is used to drive the  punch. The arms then retract back to the starting  position, while the legs return to their original position. \\ \hline
A person is balancing while  moving slowly forward. &
  The arms are extended out to the sides for balance,  while the legs are moving forward in a slow, steady pace. \\ \hline
A human walking forward. &
  The arms swing naturally at the sides of the body, while  the legs move in a forward motion, one after the other,  to propel the body forward.\\  \hline
A person is doing jumping jacks. &
  The arms and legs move in a coordinated manner, with  the arms extending out to the sides and the legs jumping  out and then back in. The arms and legs are typically  held in a straight position during the jump.\\ \hline
A person prays with both hands  and does the sign of the cross. &
  The person bows their head and brings their hands together  in front of their chest. They then make the sign of the cross  by touching their forehead, chest, and shoulders with their  right hand, and then their left hand. \\  \hline
A person bows before someone  and then waves to that person  with his right hand. &
  The person bows before someone by bending their knees  and lowering their upper body. Then, they wave to that  person with their right hand by extending their arm and  moving their hand up and down.  \\ \hline
\end{tabular}
\caption{Examples of motion captions in the datasets KIT and HumanML3D and their respective explanation generated by the Falcon 40B LLM.}
\label{tab:explanations}
\end{table*}

\subsection{Low-Level Description Generation and Encoding}

We use Falcon 40B \cite{falcon} to generate detailed explanations -- i.e., a lower-level description -- of the motions based on the high-level motion caption. This lower-level description provides a more granular explanation of the motion's components. Examples are presented in Table \ref{tab:explanations}. For instance, the low-level description of the motion "doing jumping jacks" specifies how the arms remain straight and extend outward to the sides in coordination with the legs, which move in an opening and closing motion. In the motion caption of a person performing the sign of the cross, the LLM details the sequence of arm movements, specifying how the arm touches different parts of the body. 

The low-level descriptions generated by Falcon 40B are encoded using the Sentence-BERT text encoder before training begins. Since the text encoder is fine-tuned during training, the encodings of these low-level descriptions must be updated to reflect the model's parameter changes. To minimize computational overhead, we propose updating the encodings only at the end of each epoch.

Note that only a small fraction of the low-level captions (fewer than 50) generated by Falcon 40B were verified. Verifying tens of thousands of captions would be prohibitively expensive. However, this is not unique to our method. Text-to-image models also rely on datasets containing billions of image-text pairs that are often noisy and unverified. Despite this lack of verification, text-to-image models have achieved outstanding performance.

\subsection{Database}

The high-level descriptions, along with their corresponding low-level descriptions and encoded representations, are stored in a dynamic database. This database can be enriched post-training by adding new entries, thus allowing the model to adapt and expand its understanding. This process mirrors how humans observe a motion, describe it in low-level terms (e.g., joint movements), and then reference external information to associate the low-level description with a high-level concept.


\section{Experiments and Discussion}

In this section, we outline the experimental setup, present the results of our method in comparison with previous methods on motion-to-text, and provide ablation studies to evaluate the different components of our method.

\subsection{Experimental Setup}

\begin{table*}[tb]
\centering
\begin{tabular}{lcccccc}
\hline
 &
  \multicolumn{1}{l}{\textbf{\# Words}} &
  \multicolumn{1}{l}{\textbf{\# Motions}} &
  \multicolumn{1}{l}{\textbf{\# Captions}} &
  \multicolumn{1}{l}{\textbf{\# Lemmas}} &
  \multicolumn{1}{l}{\textbf{\# Words / \# Motions}} &
  \multicolumn{1}{l}{\textbf{\# Lemmas / \# Motions}} \\ \hline
KIT       & 4,567  & 6,018  & 12,704  & 2,375 & 0.76 & 0.39 \\
HumanML3D & 15,573 & 29,227 & 89,940  & 7,090 & 0.53 & 0.24 \\
BOTH57M   & 10,885 & 3,229  & 15,132 & 5,160 & 3.37 & 1.55 \\ \hline
\end{tabular}
\caption{Statistics for the datasets utilized. A word refers to any specific form of a term (e.g., "runs," "running," or "ran"), whereas a lemma is the base form of a word that encompasses all its inflected forms (e.g., the lemma for "runs," "running," and "ran" is "run"). When the ground truth is pre-processed using lemmatization, all words are reduced to their base forms. This process decreases the vocabulary size and simplifies the language generation task. We report the ratio of the number of words or lemmas to the number of motions. This ratio measures the average occurrence of words or lemmas per motion. A higher ratio indicates that the model must learn words or lemmas from fewer motions, making the task more challenging.}
\label{tab:data_stats}
\end{table*}

\textbf{Datasets.} We utilize the KIT \cite{kitml}, HumanML3D \cite{humanml3d}, and BOTH57M \cite{both57m} datasets in our work. The KIT dataset consists of 6,018 motions and 12,704 captions, whereas the HumanML3D dataset includes 29,232 motions and 89,940 captions. Unlike KIT and HumanML3D, the BOTH57M dataset includes additional body joints to capture hand motion in detail. It consists of 3,229 motions and 15,132 captions. In all of these three datasets, each motion is annotated with at least one caption. Detailed statistics of each dataset are presented in Table \ref{tab:data_stats}. Note that these datasets only provide high-level captions. To create the low-level captions required for our method, we utilized Falcon 40B \cite{falcon}. Specifically, we prompted the large language model with the high-level captions and instructed it to simplify them into detailed descriptions focusing on the movements of the limbs and torso (see Table \ref{tab:explanations}). Prior to the training, the motion data were normalized to zero mean and unit standard deviation.

\noindent
\textbf{Evaluation Metrics.} Following previous work \cite{motiongpt, chuan2022tm2t, lamp, m2tinter}, we utilize the natural language metrics BLEU-4 \cite{bleu4}, BLEU-1 \cite{bleu4}, ROUGE-L \cite{rouge}, and CIDEr \cite{cider} to evaluate the quality of the generated captions. Higher values of these metrics indicate better motion captioning.

\noindent
\textbf{Evaluation Protocol.} We compare our method against T2MT \cite{chuan2022tm2t}, MotionGPT \cite{motiongpt}, LaMP \cite{lamp}, MotionLLM \cite{motionllm}, and M2T-Interpretable \cite{m2tinter}. All these prior methods use the same training, validation, and test splits for both the KIT and HumanML3D datasets. Here, we also utilize this split protocol. For the HumanML3D dataset, 23,384 samples are allocated for training, 1,460 for validation, and 4,383 for testing. The KIT dataset comprises 4,888 training samples, 300 validation samples, and 830 testing samples. Additional details on the split of KIT and HumanML3D datasets are provided by the authors of the datasets \cite{kitml, humanml3d}.  We use splits of similar proportions in the BOTH57M dataset: 2,583 samples for training, 162 for validation, and 484 for testing. The details of these splits -- i.e., the specific motions assigned to each split -- are available on this project's GitHub page.

T2MT and M2T-Interpretable pre-process captions by applying lemmatization, a technique that reduces inflectional forms of words to their base or root forms. While this approach simplifies the training of the language decoder, it limits the ability of these methods to generate text with verb conjugation or noun declension. This signifies a less natural language output. In contrast, MotionGPT, MotionLLM, and LaMP do not pre-process the ground truth text, as they leverage pre-trained language models that can handle complex linguistic structures directly. For a broader and fairer comparison of our method, we conduct experiments under both lemmatized and non-lemmatized settings. 

The baseline results for the KIT and HumanML3D datasets are directly extracted from their respective papers, as they follow the same evaluation protocol. However, for BOTH57M, since the baselines have not been previously trained on this dataset, we use the available source code for TM2T and MotionGPT and employ the recommended hyper-parameters to obtain results for this dataset.

\noindent
\textbf{Implementation.} The implementation was carried out using PyTorch 2.4.1 and executed on an NVIDIA A40 GPU with 48 GB of VRAM and an Intel(R) Xeon(R) Platinum 8480+ CPU. The optimal hyper-parameters were determined through a manual search process. Training was conducted for 10 epochs. The model with the best average performance (averaged across all the four aforementioned metrics) on the validation set was selected.

\noindent
\textbf{Hyper-parameters.} The learning rate for all neural networks was set to 1e-4. The parameter $k$ was found to be optimal at 2 for the KIT, 1 for BOTH57M, and 3 for HumanML3D. The patch size for the ViT motion encoder was configured as 16×32 (time × joint states) for KIT, and 32×32 for HumanML3D and BOTH57M. The parameter $c$ was set to 0.7 for KIT and BOTH57M, and 0.5 for HumanML3D. The batch size was set to 8 for all datasets.
The parameters $\lambda_1$, $\lambda_2$, and $\lambda_3$ were all set to 1. Distilled-GPT2 is utilized for both KIT and BOTH57M dataset, whereas GPT-2 is employed for HumanML3D. 

\begin{table}[tb]
\centering
\setlength{\tabcolsep}{4pt}
\begin{tabular}{lcccc}
\hline
 & \multicolumn{1}{l}{\textbf{BLEU-4}} & \multicolumn{1}{l}{\textbf{BLEU-1}} & \multicolumn{1}{l}{\textbf{ROUGE-L}} & \multicolumn{1}{l}{\textbf{CIDEr}} \\ \hline
\multicolumn{5}{c}{\textbf{KIT Dataset}}                                                            \\ \hline
T2MT    & 18.4                 & 46.7                 & 44.2                 & 79.5                 \\
M2T-INT & 24.4                 & 58.4                 & 58.3                 & 112.1                \\
HiCAM2T    & 40.4    &  80.9      & \textbf{63.3}     &  \textbf{203.0}   \\
HiCAM2T+   & \textbf{42.1}         & \textbf{82.8}          & 60.4             & 169.8       \\ \hline
\multicolumn{5}{c}{\textbf{HumanML3D Dataset}}                                                      \\ \hline
T2MT    & 22.3       & 61.7        & 49.2     & 72.5        \\
M2T-INT & 25.0                 & 69.9                 & \textbf{55.3}        & 61.6                 \\
HiCAM2T    & 26.4                 & 68.0                 & 51.4     & \textbf{79.5}        \\
HiCAM2T+   & \textbf{28.9}        & \textbf{70.6}        & 51.4                 & 76.4        \\ \hline
\multicolumn{5}{c}{\textbf{BOTH57M Dataset}}                                                        \\ \hline
T2MT    &  {8.6} &  {34.4} &  {29.8} &  {12.2} \\
HiCAM2T    &  31.1 &  63.3 &  49.5 &  \textbf{39.4} \\
HiCAM2T+   &  {\textbf{33.6}} &  {\textbf{65.4}} &  {\textbf{51.0}} &  38.9 \\ \hline
\end{tabular}
\caption{Motion caption generation performance following the evaluation protocol by \citet{chuan2022tm2t} and \citet{m2tinter} -- i.e., utilizing lemmatization as ground truth text pre-processing. Note that M2T-INT is specifically designed for the motion representation used in KIT and HumanML3D. Adapting M2T-INT to a different motion representation, such as the one employed in BOTH57M with additional body joints, is not supported without significant modifications to the method.}
\label{tab:res_With}
\end{table}

\begin{table}[tb]
\centering
\setlength{\tabcolsep}{4pt}
\begin{tabular}{lcccc}
\hline
 & \multicolumn{1}{l}{\textbf{BLEU-4}} & \multicolumn{1}{l}{\textbf{BLEU-1}} & \multicolumn{1}{l}{\textbf{ROUGE-L}} & \multicolumn{1}{l}{\textbf{CIDEr}} \\ \hline
\multicolumn{5}{c}{\textbf{HumanML3D Dataset}}                            \\ \hline
T2MT      & 7.0           & 48.9          & 38.1          & 16.8          \\
MotionGPT & 12.5          & 48.2          & 37.4          & 29.2          \\
LaMP      & 13.0          & 47.8          & 37.1          & 28.9          \\
MotionLLM      & 16.7          & 54.3          & 41.2          & 42.3          \\
HiCAM2T      & 22.4          & 61.8          & 47.0          & 44.8          \\
HiCAM2T+     & \textbf{27.8} & \textbf{67.4} & \textbf{50.4} & \textbf{54.0} \\ \hline
\multicolumn{5}{c}{\textbf{BOTH57M Dataset}}                            \\ \hline
MotionGPT & 5.9  & 43.0   &   37.1   &   2.7  \\ 
HiCAM2T      & 25.8      & 55.5         & 43.7        & 22.1      \\
HiCAM2T+     & \textbf{31.9}                                   &     \textbf{61.9}                              &      \textbf{48.3}                                &       \textbf{27.6}      \\ \hline
\end{tabular}
\caption{Motion caption generation performance following the evaluation protocol by \citet{motiongpt} and \citet{lamp} -- i.e., without ground-truth text pre-processing. }
\label{tab:res_without}
\end{table}

\subsection{Results}

Table \ref{tab:res_With} and Table \ref{tab:res_without} present the experimental results with and without ground-truth pre-processing, respectively. In both tables, our method is shown in two variants: \textbf{HiCAM2T} and \textbf{HiCAM2T+}. The difference is that the latter uses an enriched database that encompasses motion captions from both the training set and the validation set, whereas the former variant relies solely on motion captions from the training set.

With and without ground-truth pre-processing (i.e., transforming inflected words into their base or root forms), our method HiCAM2T consistently outperforms previous approaches in the literature. The performance gains are particularly notable in more challenging scenarios, such as in the absence of ground-truth pre-processing or for the BOTH57M dataset. These scenarios involve a higher ratio of words or lemmas per motion (Table \ref{tab:data_stats}), which requires the model to learn a greater number of terms from fewer motion instances. Our method achieves superior performance in these more challenging settings by leveraging explicit knowledge from a database, rather than relying on the language decoder to store terms (words or lemmas) as implicit knowledge in its learnable parameters.

We observe a notable performance improvement ranging from 1\% to 13\% in the HiCAM2T+ variant (which utilizes an expanded database) compared to the simpler HiCAM2T variant (which excludes captions from the validation set). HiCAM2T+ demonstrates superior performance in the more challenging setting when the ground truth is not pre-processed to reduce inflected forms to their lemmas.

\subsection{The Influence of $k$}

Table \ref{tab:kresults} shows that the optimal value for $k$ is relatively low, with $k=2$ for the KIT and BOTH57M datasets, and $k=3$ for HumanML3D. Lower $k$ values may indicate that the text decoder lacks sufficient information to generate accurate high-level captions. On the other hand, higher $k$ values can negatively impact performance due to two primary factors: (1) the introduction of noise, as the likelihood of retrieving irrelevant information increases, and (2) possible ambiguity arising from conflicting retrievals that confuse the model.

\begin{table}[tb]
\begin{tabular}{lcccc}
\hline
\textbf{k value} & \multicolumn{1}{l}{\textbf{BLEU-4}} & \multicolumn{1}{l}{\textbf{BLEU-1}} & \multicolumn{1}{l}{\textbf{ROUGE-L}} & \multicolumn{1}{l}{\textbf{CIDEr}} \\ \hline
\multicolumn{5}{c}{\textbf{HumanML3D Dataset}}                                                                                                                           \\ \hline
k = 1            &     25.8                                &     66.3                                &      49.5                                &      52.5                              \\
k = 2            &    25.0                                 &  65.4                                    &     48.5                                 &                         52.0          \\
k = 3            & \textbf{27.8}                       & \textbf{67.4}                       & \textbf{50.4}                        & \textbf{54.0}                      \\
k = 4            & 18.2                           & 52.2                          & 41.9                  & 28.6                          \\ \hline
\multicolumn{5}{c}{\textbf{KIT Dataset}}                                                                                                                                 \\ \hline
k = 1          &   23.6                              &   52.6                                  &       50.9                           &        82.9                                     \\
k = 2            & 25.3                       & \textbf{54.6}                       & \textbf{51.9}                        & \textbf{90.2}                      \\
k = 3            &  23.1                                   &   50.5                                  &            51.6                          &         82.2                           \\
k = 4            & \textbf{25.4}                           & 52.0                           & 48.5                            & 67.5                          \\ \hline
\multicolumn{5}{c}{\textbf{BOTH57M Dataset}}                                                                                                                             \\ \hline
k = 1            &   \textbf{31.9}                                   &     \textbf{61.9}                              &      \textbf{48.3}                                &       \textbf{27.6}                             \\
k = 2            & 28.1                       & 58.0                       & 45.6                        & 26.8                      \\
k = 3            &   23.3                                  &           52.5                          &    40.7                                  &         17.2                           \\
k = 4            & 21.8         &  49.1                       & 37.8                    & 15.2                  \\ \hline
\end{tabular}
\caption{Motion caption generation performance for different values of the hyper-parameter $k$, without ground-truth pre-processing (lemmatization).}
\label{tab:kresults}
\end{table}

\begin{table}[tb]
\centering
\setlength{\tabcolsep}{2.5pt}
\begin{tabular}{lcccc}
\hline
 & \multicolumn{1}{l}{\textbf{BLEU-4}} & \multicolumn{1}{l}{\textbf{BLEU-1}} & \multicolumn{1}{l}{\textbf{ROUGE-L}} & \multicolumn{1}{l}{\textbf{CIDEr}} \\ \hline
\multicolumn{5}{c}{\textbf{HumanML3D Dataset}}                                                                                                \\ \hline
Base                                                                          & 7.4           & 47.3          & 31.1          & 7.00          \\
\begin{tabular}[c]{@{}l@{}}Frozen $TD$ and \\ no $\mathcal{L}_2$\end{tabular} & 9.2           & 38.9          & 30.7          & 12.3          \\
No $\mathcal{L}_2$                                                            & 23.1          & 63.0          & 47.4          & 46.9          \\
Top-1 directly  &       12.4                                                  &     52.6      &  37.1     &  26.5    \\
Complete                                                                      & \textbf{27.8} & \textbf{67.4} & \textbf{50.4} & \textbf{54.0} \\ \hline
\multicolumn{5}{c}{\textbf{KIT Dataset}}                                                                                                      \\ \hline
Base  & 14.6          & 43.5          & 40.9          & 40.1          \\
\begin{tabular}[c]{@{}l@{}}Frozen $TD$ and \\ no $\mathcal{L}_2$\end{tabular} & 21.5          & 50.7          & 49.0          & 69.2          \\
No $\mathcal{L}_2$                                                            & 22.2          & 50.1          & 49.8          & 82.0          \\
Top-1 directly  &    15.4                                                &  46.2         &  42.2     &  43.5    \\
Complete                                                                      & \textbf{25.3} & \textbf{54.6} & \textbf{51.9} & \textbf{90.2} \\ \hline
\multicolumn{5}{c}{\textbf{BOTH57M Dataset}}                                                                                                  \\ \hline
Base                                                                          &         14.6      &        48.6       &    34.8           &    8.2         \\
\begin{tabular}[c]{@{}l@{}}Frozen $TD$ and \\ no $\mathcal{L}_2$\end{tabular} &   18.5        &    46.8   &  35.3  &    9.7    \\
No $\mathcal{L}_2$     &  25.9                                                      &    56.6           &        44.5       &       21.0                      \\
Top-1 directly  &       18.0                                                 &    52.9       &     37.2  &  12.6    \\
Complete    & \textbf{31.9 }     & \textbf{61.9 }     & \textbf{48.3}     & \textbf{27.6}     \\ \hline
\end{tabular}
\caption{Ablation studies on HumanML3D, KIT, and BOTH57M datasets without ground-truth pre-processing.}
\label{tab:ablation}
\vspace{-0.5cm}
\end{table}

\begin{figure}[tb]
    \centering
    \includegraphics[width=0.96\linewidth]{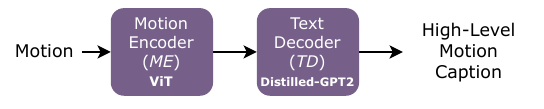}
    \caption{Base variant in the ablation studies. }
    \label{fig:base}
\end{figure}

\subsection{Ablation Studies}

To evaluate the individual components of our method, we conducted ablation studies with five variants. The first variant, \textbf{Complete}, represents the full method as described in the methodology section. The second variant differs from the complete variant by utilizing the top-1 retrieved high-level caption directly as the final output, bypassing the combined processing of all retrieved high-level with the motion features in the text decoder. The third variant excludes the contrastive loss term $\mathcal{L}_2$. The fourth variant freezes the text decoder during training in addition to excluding $\mathcal{L}_2$.  Finally, the fifth variant, referred to as \textbf{Base}, entirely departs from the architecture we developed. In the \textbf{Base} variant, the setup consists solely of the motion encoder and a text decoder. Both the motion encoder and the text decoder are trained, and the output of the motion encoder is passed directly to the text decoder to generate high-level descriptions (Fig. \ref{fig:base}).

Table \ref{tab:ablation} presents the results of the ablation studies. Note that for these studies we chose the more complex scenario without ground-truth pre-processing  (i.e., lemmatization). The results demonstrate that generating high-level captions directly from motion (\textbf{Base} variant) leads to significant performance degradation. This approach produces results similar to T2MT and lower than MotionGPT, LaMP, and M2T-INT, which employ alternative strategies for motion encoding and language modeling to enhance performance while directly mapping from motion to high-level captions. We also observe that even the simplest hierarchical approach (excluding both text decoder training and the $\mathcal{L}_2$ loss term) provides a considerable performance improvement. Further gains are achieved when fine-tuning both the text decoder and the text encoder with the $\mathcal{L}_2$ loss term.

The performance of directly utilizing the top-1 retrieved high-level caption as the final output reflects the effectiveness of the retrieval mechanism when $k = 1$. We observe a significant performance drop compared to the complete method. This emphasizes that the final step of the method successfully integrates information from all retrieved captions and the motion features to generate a meaningful output.

\subsection{Limitations and Future Work}

\textbf{Ambiguity.} Low-level and high-level captions do not always have a one-to-one correspondence. A single low-level caption can describe multiple distinct motions, which may lead to ambiguity. For example, the low-level caption ``the subject raises their right arm, bends the elbow slightly, and moves their hand side to side in a repeated motion." could correspond to both ``the person waves energetically to greet a friend." and ``the person quickly swats at an insect in the air." The correct interpretation depends on the context in which the motion occurs.

However, this ambiguity challenge is not unique to our work but is a fundamental limitation of all motion-to-text models. This is because a single motion can also correspond to multiple high-level captions. In other words, there is no strict one-to-one mapping between low-level and high-level captions, nor between high-level captions and motion. Hence, the context in which a motion occurs plays a crucial role in determining its most appropriate description.

As future work, we aim to reduce the ambiguity between low-level and high-level captions, as well as between high-level captions and motions, by conditioning the output on context. We aim at achieving this without relying on extensive data collection of motion-text pairs across multiple contexts.

\textbf{Noisy Low-Level Captions.} As previously mentioned, low-level captions generated by LLMs may contain noise, and verifying thousands of them manually is impractical. As future work, we aim to develop methods for automatically filtering noisy low-level captions and conditioning the low-level caption generation on images of the motion. Temporal information can be also incorporated into the generation process by providing as input the body joints exhibit the most movement and at what speed. Additionally, as future work, the retrieval mechanism can be enhanced by conditioning it also on the input motion.

\subsection{Granularity-Adaptive Captions} Current motion-to-text models exclusively focus on generating high-level captions as their final output, as these are more human-friendly and concise. In other words, they do not require the viewer to mentally piece together a sequence of actions before recognizing the movement, making them less intuitive. However, in some cases, a caption with more granular details may be more beneficial. For instance, when a detailed analysis of specific actions or movements is required -- such as in sports analysis or medical diagnostics -- mid-level descriptions can offer a deeper understanding by capturing important nuances that high-level captions might miss. As future work, we aim to develop methods for adjusting the granularity of motion captions based on the use case.


\section{Conclusions}

In this paper, we propose a novel motion-to-text framework for time-series data (encompassing body joint values such as angles and velocities over time) that abstains from directly mapping motion to high-level descriptions. Instead, our hierarchical approach first generates low-level descriptions that detail the movements of body joints, which are then used to produce high-level captions. This method achieves significant performance improvements over existing approaches.



\bibliography{references}

\begin{thebibliography}{25}
\providecommand{\natexlab}[1]{#1}
\providecommand{\url}[1]{\texttt{#1}}
\expandafter\ifx\csname urlstyle\endcsname\relax
  \providecommand{\doi}[1]{doi: #1}\else
  \providecommand{\doi}{doi: \begingroup \urlstyle{rm}\Url}\fi

\bibitem[Almazrouei et~al.(2023)Almazrouei, Alobeidli, Alshamsi, Cappelli, Cojocaru, Debbah, Étienne Goffinet, Hesslow, Launay, Malartic, Mazzotta, Noune, Pannier, and Penedo]{falcon}
Almazrouei, E., Alobeidli, H., Alshamsi, A., Cappelli, A., Cojocaru, R., Debbah, M., Étienne Goffinet, Hesslow, D., Launay, J., Malartic, Q., Mazzotta, D., Noune, B., Pannier, B., and Penedo, G.
\newblock The falcon series of open language models, 2023.
\newblock URL \url{https://arxiv.org/abs/2311.16867}.

\bibitem[Chen et~al.(2024)Chen, Lu, Zeng, Zhang, Wang, Zhang, and Zhang]{motionllm}
Chen, L.-H., Lu, S., Zeng, A., Zhang, H., Wang, B., Zhang, R., and Zhang, L.
\newblock Motionllm: Understanding human behaviors from human motions and videos, 2024.
\newblock URL \url{https://arxiv.org/abs/2405.20340}.

\bibitem[Dosovitskiy et~al.(2020)Dosovitskiy, Beyer, Kolesnikov, Weissenborn, Zhai, Unterthiner, Dehghani, Minderer, Heigold, Gelly, Uszkoreit, and Houlsby]{vit}
Dosovitskiy, A., Beyer, L., Kolesnikov, A., Weissenborn, D., Zhai, X., Unterthiner, T., Dehghani, M., Minderer, M., Heigold, G., Gelly, S., Uszkoreit, J., and Houlsby, N.
\newblock An image is worth 16x16 words: Transformers for image recognition at scale.
\newblock \emph{CoRR}, abs/2010.11929, 2020.
\newblock URL \url{https://arxiv.org/abs/2010.11929}.

\bibitem[Guo et~al.(2022{\natexlab{a}})Guo, Zou, Zuo, Wang, Ji, Li, and Cheng]{humanml3d}
Guo, C., Zou, S., Zuo, X., Wang, S., Ji, W., Li, X., and Cheng, L.
\newblock Generating diverse and natural 3d human motions from text.
\newblock In \emph{Proceedings of the IEEE/CVF Conference on Computer Vision and Pattern Recognition (CVPR)}, pp.\  5152--5161, June 2022{\natexlab{a}}.

\bibitem[Guo et~al.(2022{\natexlab{b}})Guo, Zuo, Wang, and Cheng]{chuan2022tm2t}
Guo, C., Zuo, X., Wang, S., and Cheng, L.
\newblock Tm2t: Stochastic and tokenized modeling for the reciprocal generation of 3d human motions and texts.
\newblock In \emph{ECCV}, 2022{\natexlab{b}}.

\bibitem[Guo et~al.(2023)Guo, Mu, Javed, Wang, and Cheng]{momask}
Guo, C., Mu, Y., Javed, M.~G., Wang, S., and Cheng, L.
\newblock Momask: Generative masked modeling of 3d human motions.
\newblock 2023.

\bibitem[Hu et~al.(2021)Hu, Shen, Wallis, Allen{-}Zhu, Li, Wang, and Chen]{lora}
Hu, E.~J., Shen, Y., Wallis, P., Allen{-}Zhu, Z., Li, Y., Wang, S., and Chen, W.
\newblock Lora: Low-rank adaptation of large language models.
\newblock \emph{CoRR}, abs/2106.09685, 2021.
\newblock URL \url{https://arxiv.org/abs/2106.09685}.

\bibitem[Jiang et~al.(2024)Jiang, Chen, Liu, Yu, Yu, and Chen]{motiongpt}
Jiang, B., Chen, X., Liu, W., Yu, J., Yu, G., and Chen, T.
\newblock Motiongpt: human motion as a foreign language.
\newblock In \emph{Proceedings of the 37th International Conference on Neural Information Processing Systems}, NIPS '23, Red Hook, NY, USA, 2024. Curran Associates Inc.

\bibitem[Lamiae et~al.(2023)Lamiae, Hicham, Fatiha, Mohammed, and Hajoub]{10.1145/3607720.3607746}
Lamiae, E., Hicham, G.~T., Fatiha, E., Mohammed, B., and Hajoub, M.~W.
\newblock Patient smart home monitoring using vision neural network transformers.
\newblock In \emph{Proceedings of the 6th International Conference on Networking, Intelligent Systems \& Security}, NISS '23, New York, NY, USA, 2023. Association for Computing Machinery.
\newblock ISBN 9798400700194.
\newblock \doi{10.1145/3607720.3607746}.
\newblock URL \url{https://doi.org/10.1145/3607720.3607746}.

\bibitem[Leite et~al.(2024)Leite, Mauranen, Zhanabatyrova, and Xiao]{leite2024transformerbasedapproachessensorbasedhuman}
Leite, C.~S., Mauranen, H., Zhanabatyrova, A., and Xiao, Y.
\newblock Transformer-based approaches for sensor-based human activity recognition: Opportunities and challenges, 2024.
\newblock URL \url{https://arxiv.org/abs/2410.13605}.

\bibitem[Li et~al.(2024)Li, Yuan, He, Qiu, Zhu, Gu, Shen, Dong, Dong, and Yang]{lamp}
Li, Z., Yuan, W., He, Y., Qiu, L., Zhu, S., Gu, X., Shen, W., Dong, Y., Dong, Z., and Yang, L.~T.
\newblock Lamp: Language-motion pretraining for motion generation, retrieval, and captioning, 2024.
\newblock URL \url{https://arxiv.org/abs/2410.07093}.

\bibitem[Lin(2004)]{rouge}
Lin, C.-Y.
\newblock {ROUGE}: A package for automatic evaluation of summaries.
\newblock In \emph{Text Summarization Branches Out}, pp.\  74--81, Barcelona, Spain, July 2004. Association for Computational Linguistics.
\newblock URL \url{https://aclanthology.org/W04-1013}.

\bibitem[Papineni et~al.(2002)Papineni, Roukos, Ward, and Zhu]{bleu4}
Papineni, K., Roukos, S., Ward, T., and Zhu, W.-J.
\newblock Bleu: a method for automatic evaluation of machine translation.
\newblock In \emph{Proceedings of the 40th Annual Meeting on Association for Computational Linguistics}, ACL '02, pp.\  311–318, USA, 2002. Association for Computational Linguistics.
\newblock \doi{10.3115/1073083.1073135}.
\newblock URL \url{https://doi.org/10.3115/1073083.1073135}.

\bibitem[Plappert et~al.(2016)Plappert, Mandery, and Asfour]{kitml}
Plappert, M., Mandery, C., and Asfour, T.
\newblock The {KIT} motion-language dataset.
\newblock \emph{Big Data}, 4\penalty0 (4):\penalty0 236--252, dec 2016.
\newblock \doi{10.1089/big.2016.0028}.
\newblock URL \url{http://dx.doi.org/10.1089/big.2016.0028}.

\bibitem[Radford et~al.(2019)Radford, Wu, Child, Luan, Amodei, and Sutskever]{radford2019language}
Radford, A., Wu, J., Child, R., Luan, D., Amodei, D., and Sutskever, I.
\newblock Language models are unsupervised multitask learners.
\newblock 2019.

\bibitem[Radouane et~al.(2023)Radouane, Tchechmedjiev, Ranwez, and Lagarde]{m2tinter}
Radouane, K., Tchechmedjiev, A., Ranwez, S., and Lagarde, J.
\newblock {Guided Attention for Interpretable Motion Captioning}.
\newblock working paper or preprint, October 2023.
\newblock URL \url{https://imt-mines-ales.hal.science/hal-04251363}.

\bibitem[Raffel et~al.(2020)Raffel, Shazeer, Roberts, Lee, Narang, Matena, Zhou, Li, and Liu]{t5}
Raffel, C., Shazeer, N., Roberts, A., Lee, K., Narang, S., Matena, M., Zhou, Y., Li, W., and Liu, P.~J.
\newblock Exploring the limits of transfer learning with a unified text-to-text transformer.
\newblock \emph{J. Mach. Learn. Res.}, 21\penalty0 (1), January 2020.
\newblock ISSN 1532-4435.

\bibitem[Reimers \& Gurevych(2019)Reimers and Gurevych]{sentence-bert}
Reimers, N. and Gurevych, I.
\newblock Sentence-bert: Sentence embeddings using siamese bert-networks.
\newblock In \emph{Proceedings of the 2019 Conference on Empirical Methods in Natural Language Processing}. Association for Computational Linguistics, 11 2019.
\newblock URL \url{https://arxiv.org/abs/1908.10084}.

\bibitem[Safiya \& Pandian(2024)Safiya and Pandian]{Safiya2024}
Safiya, K.~M. and Pandian, R.
\newblock A real-time image captioning framework using computer vision to help the visually impaired.
\newblock \emph{Multimedia Tools and Applications}, 83\penalty0 (20):\penalty0 59413--59438, Jun 2024.
\newblock ISSN 1573-7721.
\newblock \doi{10.1007/s11042-023-17849-7}.
\newblock URL \url{https://doi.org/10.1007/s11042-023-17849-7}.

\bibitem[Sanh et~al.(2019)Sanh, Debut, Chaumond, and Wolf]{sanh2019distilbert}
Sanh, V., Debut, L., Chaumond, J., and Wolf, T.
\newblock Distilbert, a distilled version of bert: smaller, faster, cheaper and lighter.
\newblock In \emph{NeurIPS EMC2 Workshop}, 2019.

\bibitem[van~den Oord et~al.(2017)van~den Oord, Vinyals, and Kavukcuoglu]{vqvae}
van~den Oord, A., Vinyals, O., and Kavukcuoglu, K.
\newblock Neural discrete representation learning.
\newblock \emph{CoRR}, abs/1711.00937, 2017.
\newblock URL \url{http://arxiv.org/abs/1711.00937}.

\bibitem[Vedantam et~al.(2015)Vedantam, Zitnick, and Parikh]{cider}
Vedantam, R., Zitnick, C.~L., and Parikh, D.
\newblock Cider: Consensus-based image description evaluation, 2015.
\newblock URL \url{https://arxiv.org/abs/1411.5726}.

\bibitem[Zhang et~al.(2023)Zhang, Zhang, Cun, Huang, Zhang, Zhao, Lu, and Shen]{t2mgpt}
Zhang, J., Zhang, Y., Cun, X., Huang, S., Zhang, Y., Zhao, H., Lu, H., and Shen, X.
\newblock T2m-gpt: Generating human motion from textual descriptions with discrete representations.
\newblock In \emph{Proceedings of the IEEE/CVF Conference on Computer Vision and Pattern Recognition (CVPR)}, 2023.

\bibitem[Zhang et~al.(2022)Zhang, Roller, Goyal, Artetxe, Chen, Chen, Dewan, Diab, Li, Lin, Mihaylov, Ott, Shleifer, Shuster, Simig, Koura, Sridhar, Wang, and Zettlemoyer]{opt27b}
Zhang, S., Roller, S., Goyal, N., Artetxe, M., Chen, M., Chen, S., Dewan, C., Diab, M., Li, X., Lin, X.~V., Mihaylov, T., Ott, M., Shleifer, S., Shuster, K., Simig, D., Koura, P.~S., Sridhar, A., Wang, T., and Zettlemoyer, L.
\newblock Opt: Open pre-trained transformer language models, 2022.
\newblock URL \url{https://arxiv.org/abs/2205.01068}.

\bibitem[Zhang et~al.(2024)Zhang, Huang, Zhou, Zhang, Yu, Wang, and Xu]{both57m}
Zhang, W., Huang, M., Zhou, Y., Zhang, J., Yu, J., Wang, J., and Xu, L.
\newblock Both2hands: Inferring 3d hands from both text prompts and body dynamics.
\newblock In \emph{2024 IEEE/CVF Conference on Computer Vision and Pattern Recognition (CVPR)}, pp.\  2393–2404. IEEE, June 2024.
\newblock \doi{10.1109/cvpr52733.2024.00232}.
\newblock URL \url{http://dx.doi.org/10.1109/CVPR52733.2024.00232}.

\end{thebibliography}
\bibliographystyle{icml2025}

\end{document}